\documentclass{article}

\usepackage[preprint]{icml2019}
\usepackage{amssymb,amsmath}
\usepackage{multirow}
\usepackage{times}
\usepackage{graphicx}
\usepackage{placeins}
\usepackage[table,xcdraw]{xcolor}
\usepackage{algorithm,algorithmic}
\usepackage{multicol}
\usepackage{amsthm}
\usepackage{colortbl}
\usepackage{wrapfig}
\DeclareMathOperator*{\argmax}{arg\,max}

\usepackage{calc} %
\usepackage{wrapfig}

\usepackage{floatflt}
\usepackage{hyperref}%

\definecolor{mydarkblue}{rgb}{0,0.08,0.45}

\hypersetup{ %
    pdftitle={Scaling Matters in Deep Structured-Prediction Models},
    pdfauthor={Aleksandr Shevchenko, Anton Osokin},
    pdfsubject={Scaling Matters in Deep Structured-Prediction Models},
    pdfkeywords={structured prediction, deep learning},
    pdfborder=0 0 0,
    pdfpagemode=UseNone,
    colorlinks=true,
    linkcolor=mydarkblue,
    citecolor=mydarkblue,
    filecolor=mydarkblue,
    urlcolor=mydarkblue,
    pdfview=FitH
}

\begin{document}
\twocolumn[
\icmltitle{Scaling Matters in Deep Structured-Prediction Models}
\icmlsetsymbol{equal}{*}

\begin{icmlauthorlist}
\icmlauthor{Aleksandr Shevchenko}{to}
\icmlauthor{Anton Osokin}{to}
\end{icmlauthorlist}

\icmlaffiliation{to}{Samsung-HSE Laboratory, National Research University Higher School of Economics, Moscow, Russia}

\icmlcorrespondingauthor{Aleksandr Shevchenko}{ashevchenko@hse.ru}

\vskip 0.3in
]
\printAffiliationsAndNotice{}

\begin{abstract}
Deep structured-prediction energy-based models combine the expressive power of learned representations and the ability of embedding knowledge about the task at hand into the system. A common way to learn parameters of such models consists in a multistage procedure where different combinations of components are trained at different stages. The joint end-to-end training of the whole system is then done as the last fine-tuning stage. This multistage approach is time-consuming and cumbersome as it requires multiple  runs until convergence and multiple rounds of hyperparameter tuning. From this point of view, it is beneficial to start the joint training procedure from the beginning. However, such approaches often unexpectedly fail and deliver results worse than the multistage ones. In this paper, we hypothesize that one reason for joint training of deep energy-based models to fail is the incorrect relative normalization of different components in the energy function. We propose online and offline scaling algorithms that fix the joint training and demonstrate their efficacy on three different tasks.
\end{abstract}

\section{Introduction}

In structured prediction, the goal is to make multiple highly-correlated predictions jointly in a coordinated way~\citep{nowozin2011structured,smith2011linguistic}.
As an example of a structured-prediction task consider optical character recognition (OCR) where the main goal is to recognize a word given the sequence of images of hand-written letters.
One possible approach to this task is to train a classifier (can be, e.g., an MNIST-size convolutional neural network, CNN \citep{lecun1995convolutional}) that separately predicts each letter given its image (we will refer to such classifiers as unary predictors).
However, a unary predictor does not take into account the global word structure, i.e., which combinations of letters are likely to occur together.
The information about the word structure can be either taken into account by enriching the features fed into the unary predictor with information about other letter images or by explicitly modeling dependencies between letters.
We associate the models doing the latter with structured prediction (possibly together with the former).
On the OCR dataset collected by~\citet{taskar2003max}, sole unary predictors usually give 8-12\% letter recognition error and structured-prediction models can decrease the error to 1-3\%  \citep{perez07cgm,searnn2018leblond}.

One popular approach to build structured-prediction systems is the framework of undirected probabilistic graphical models~\citep{wainwright2008} or closely related energy-based models~\citep{lecun2006tutorial}.
As of today, it is common to use structured models with parts parameterized by neural networks.
Such models give competitive results in many applications and are often called deep structured-prediction models (DSPM).

To train DSPMs, one typically needs to do inference of some form at the training time and combine it with back-propagation and stochastic optimization.
Modern neural network libraries (e.g., PyTorch \citep{paszke2017automatic} and TensorFlow \citep{tensorflow2015-whitepaper}) offer powerful automatic differentiation tools that allow to compute gradients of many inference schemes, but training such models is often difficult.
A common simplifications consists in training different system components separately, which provides a good initialization for the joint fine-tuning.
We refer to this approach as \emph{stage training} to contrast it to the \emph{joint training}, which amounts to a single stage that learns all the parameters.
Stage training is time-consuming as the procedure requires multiple runs of stochastic optimization until convergence and separate hyperparameter tuning for each stage.
The process also lacks stability due to the stage switching.
In the literature, there is contradicting evidence whether the stage or joint training procedure works better (see~Section~\ref{sec:relatedworks}).

\begin{figure*}[t]
\begin{tabular}{@{}c@{\!\!\!\!\!}c@{\!\!\!\!\!}c@{}}
\includegraphics[width=0.35\textwidth, clip=true, trim=6mm 0mm 6mm 0mm]{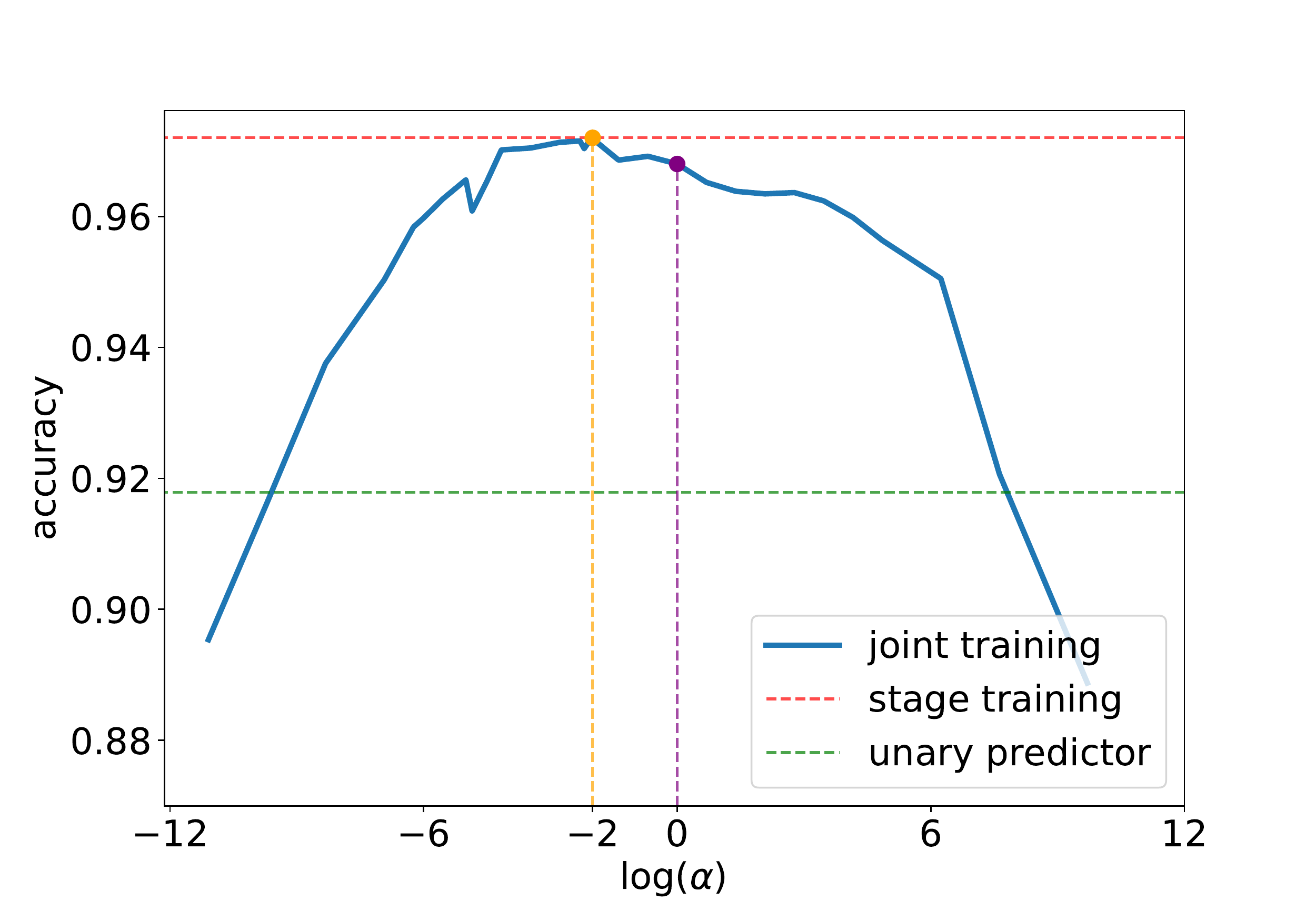}
&
\includegraphics[width=0.35\textwidth, clip=true, trim=6mm 0mm 6mm 0mm]{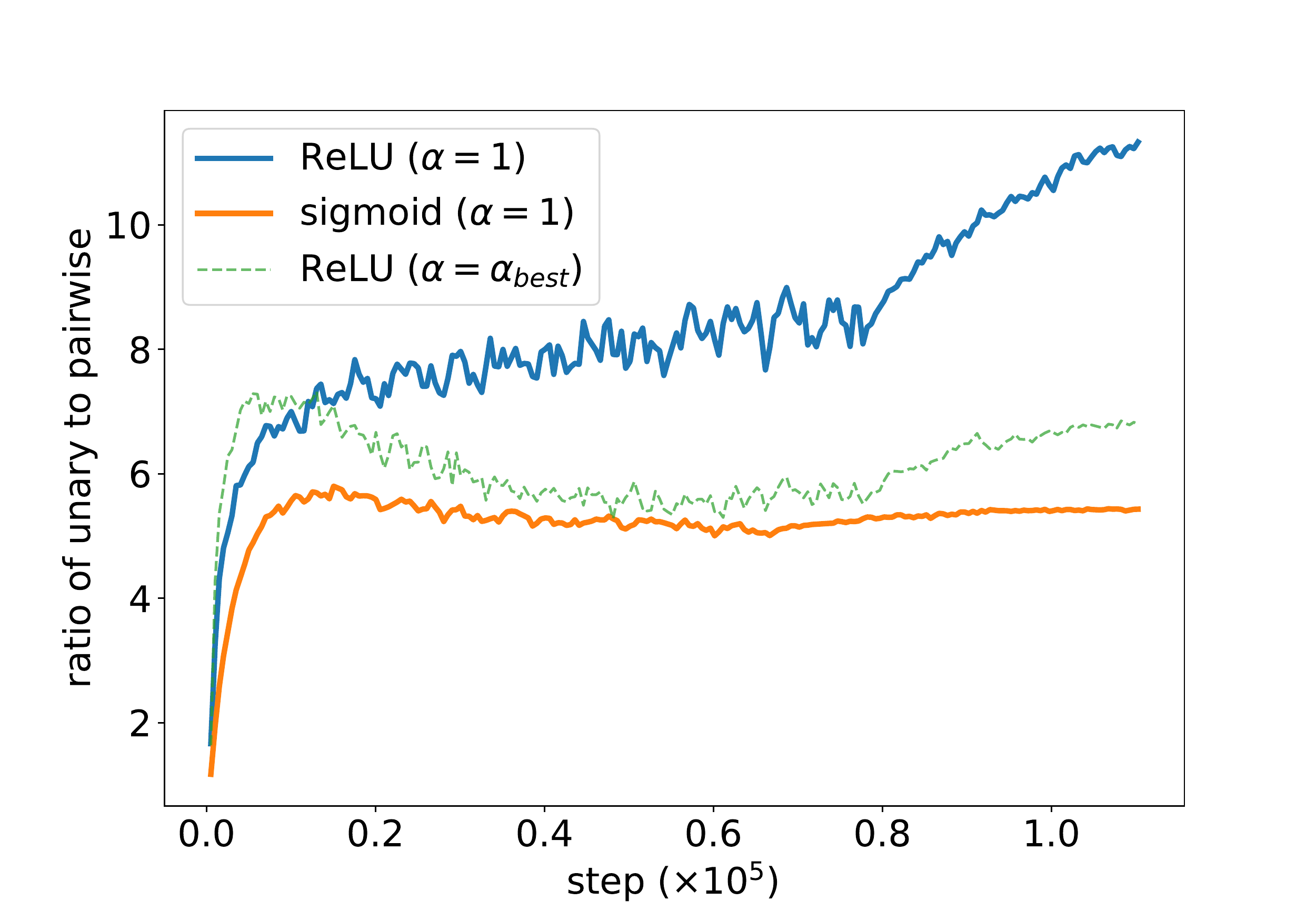}
&
\includegraphics[width=0.35\textwidth, clip=true, trim=6mm 0mm 6mm 0mm]{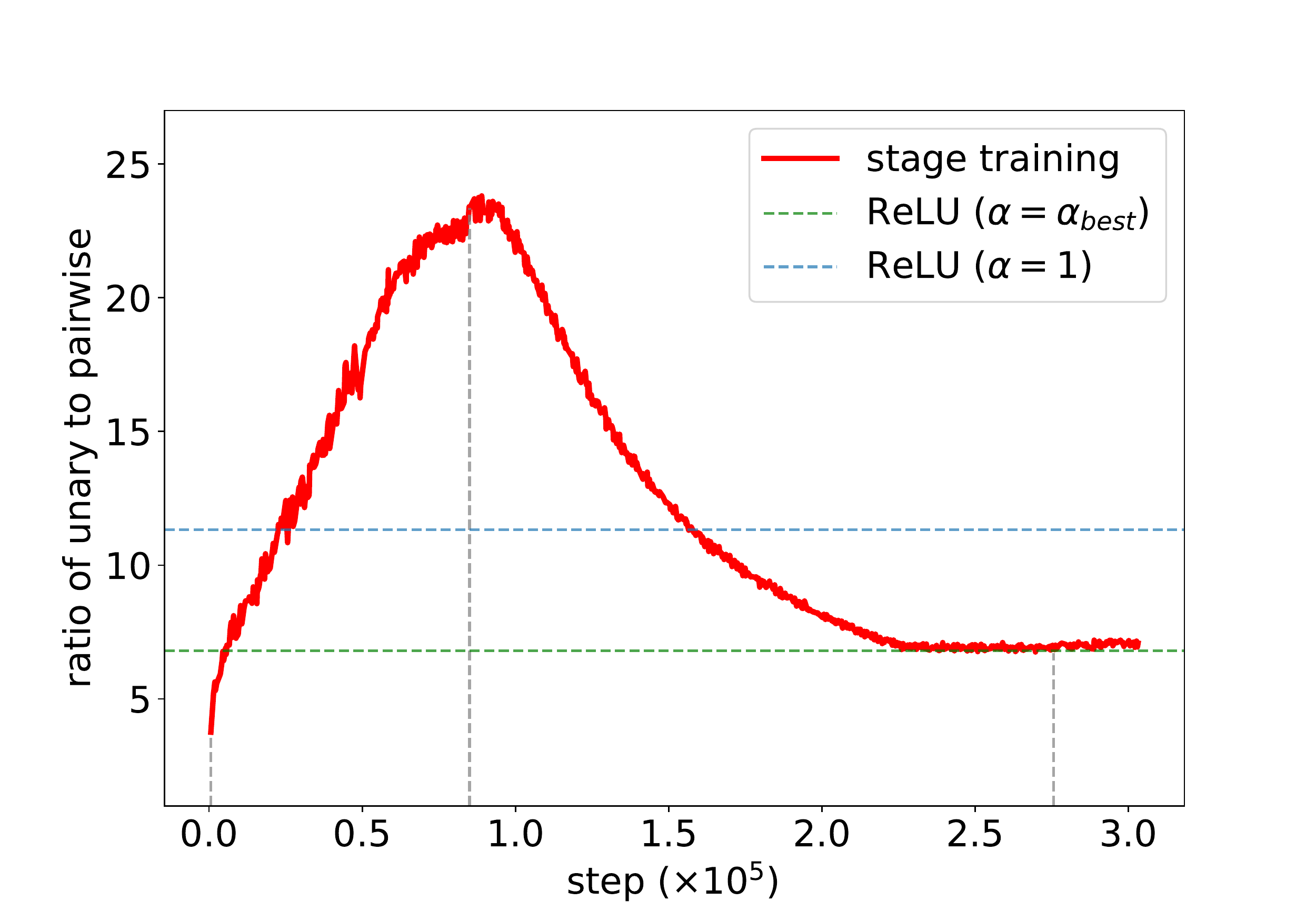}
\\[-1mm]
(a) & (b) & (c)
\end{tabular}
\vspace{-5mm}\caption{\label{motiv} OCR case study, part 2. Plot~(a) shows the dependence of trained model performance on the constant~$\alpha$ (the scaling factor of the unary potentials). The magenta vertical line corresponds to the initial setting with no scaling applied, i.e., $\alpha=1$, the yellow line corresponds to the best value of~$\alpha$, which is around $0.25$.
Plot~(b) presents the dynamics of the ratio of the unary to pairwise potentials during the joint training for the models with the different values of~$\alpha$ and different activations at the top of the net defining the unary potentials.
Plot~(c) presents the analogous dynamics for the stage training (for the first stage, we show the ratio of the unary potentials to the initialization of the pairwise ones).
The vertical dashed lines indicate the moments of stage switching.
The horizontal dashed lines indicate the quantities we compare the curve to.
} 
\end{figure*}

\textbf{Contributions.}
In this paper, we experimentally compare the stage and joint procedures and provide evidence (see Section~\ref{sec:motivation} for the OCR case study) that the main reason for joint procedure to fail in some cases\footnote{Please note that we do not consider the cases when the joint training procedures fails because of the lack of fully labeled data or some engineering reasons such as GPU memory limits.} is an improper relative scaling of different model components.
We propose the online and offline algorithms to choose scaling that fixes the joint training (see Section~\ref{sec:scaling}).
To validate these methods, we apply them to the two particular types of DSPM, namely linear chain conditional random field, LCCRF \citep{lafferty2001conditional}, with application to OCR and text chunking, and Gaussian conditional random field~\citep{GAUSS} with application to the task of image segmentation (see Section~\ref{sec:models} for a review of these models with the inference and training approaches).
We present the experimental results in Section~\ref{sec:experiments} and conclude in Section~\ref{sec:conclusion}.

\section{Motivation: OCR Case Study\label{sec:motivation}} 
To showcase the issues related to joint vs.\ stage training choice, we study a simple system for the task of optical character recognition (OCR).
We use the linear chain CRF model where the unary potentials are provided by the LeNet5-like CNN~\citep{lecun1998gradient} and the pairwise potentials are parameterized by a $26 \times 26$ matrix.
As the training objective we try the standard cross-entropy loss defined on the unary marginal distributions, structured SVM loss and log-likelihood.
See Sections~\ref{sec:models} and~\ref{sec:experiments} for details.

The joint training procedure consists in minimizing the objective with a stochastic optimization algorithm where at each iteration the gradient is obtained by automatically differentiating the inference algorithm (the sum-product or max-sum message passing).
The stage training procedure consists in the following three stages: train the unary predictor to classify the individual letter images independently, learn the matrix defining the pairwise potentials by minimizing the objective w.r.t.\ it (without back-propagating the gradients into the unary predictor), fine-tune all the components by minimizing the objective w.r.t.\ all the parameters jointly.

Table~\ref{tablemotiveocr} reports the comparison of the three different objectives for both stage and joint training procedures, where we observe that the stage training procedure is better for all objectives.
If we tweak the model slightly and change the ReLU activation of the last linear layer of the unary network to sigmoid, then the stage and joint procedure deliver results that are identical up to noise.
The motivation for our study is to find out why the models that are so close exhibit different behaviour.
We have seen evidence of such differences in a few related works (see Section~\ref{sec:relatedworks}) and those could even be the reason for the joint training procedure to perform worse than the sole unary predictors.

{\renewcommand{\arraystretch}{1.24}
\begin{table}[t]
\begin{center}
\begin{tabular}{c|c|c}
\textbf{OBJECTIVE} & \textbf{STAGE}& \textbf{JOINT} \\
\hline
Cross-entropy & \textbf{97.2} $\pm$ 0.1  & 96.5 $\pm$ 0.2 \\
Structured SVM &   \textbf{97.0} $\pm$ 0.3  & 96.4 $\pm$ 0.4 \\
Log-likelihood & \textbf{97.2} $\pm$ 0.1  & 96.5 $\pm$ 0.3 \\
\end{tabular}
\end{center}
\caption{OCR case study, part 1. Stage denotes the stage training procedure and Joint stands for the joint training.
\label{tablemotiveocr}}
\end{table}
}

We now illustrate our hypothesis that the main reason for the joint training to underperform is the relative scaling of the unary and pairwise potentials in the LCCRF model.
We change the parameterization of the unary potentials by simply multiplying them by a the positive constant~$\alpha$ and check how this change affects the performance of the model after training.
In principle, this constant~$\alpha$ multiplies the output of a linear layer of a neural network so the model parametrization might compensate for it. 
However, in the actual training runs this constant significantly affects the performance due to the fact that the training procedure optimization algorithm is not scale invariant (even if we use algorithms that try to approximate scale invariance, e.g., Adam~\citep{Adam}).

Figure~\ref{motiv}a illustrates the dependence of the final model performance (trained with the cross-entropy loss) on the constant~$\alpha$.
As the two baselines we show the accuracy of the unary predictors only and of the model trained by the stage procedure.
Please note that when $\alpha$ is too large or small the training fails, and there exists a sweet spot $\alpha_{\text{best}} \neq 1$ at which the joint training matches the stage training.

Furthermore, Figure~\ref{motiv}b shows the dynamics of the ratio of the unary and pairwise potentials\footnote{By the ratio of the unary and pairwise potentials, we mean the ratio of the average absolute values of the matrices defining the potentials. See Section~\ref{lcrf} for the details.}
at the three training runs:
the model trained with $\alpha = 1$, where the joint training underperforms;
same model with $\alpha = \alpha_{\text{best}}$, where the joint training succeeds. As another point of comparison we show the ratio for the model with the last ReLU non-linearity substituted with sigmoid, for which the joint training succeeds with $\alpha = 1$.
We observe that when the joint training underperforms the ratio of the unary to pairwise potentials diverges, but stabilizes otherwise.
As additional baseline, we show the behaviour of the ratio at the stage training procedure (Figure~\ref{motiv}c).
We observe that at the first stage the ratio grows since only unary potentials are trained, at the second stage the ratio goes down to the ``good'' level and is stable at the third stage.
Please also note that the stage training procedure needs significantly more iterations than the joint one.

\section{Related Works\label{sec:relatedworks}}

\textbf{Deep Structured-Prediction Models.} First, we review the works that combine neural networks with energy-based structured models and pay attention to whether they use joint or stage training.
The idea of combining networks and energy-based models was well known in the 90s, for example,
\citet{bottou97,lecun1998gradient} introduced graph transformers (a combination of hidden Markov models and neural networks) that were trained end-to-end and deployed as systems of hand-written character recognition.

More recently, \citet{tompson2014joint,jaderberg2014text,vu2015context,chen2015learning} used the stage training approach (with the end-to-end fine-tuning at the final stage) for the computer vision tasks of human pose estimation, free text recognition, multiple object detection, image tagging, respectively.
\citet{yang2016multi,chen2017improving} used the joint training approach for the BiLSTM-CRF models that combine neural networks techniques (word embeddings and recurrent nets with LSTM units) with LCCRF model on top for a variety of NLP tasks.

Recently, the DenseCRF model \citep{krahenbuhl2011efficient} received quite a bit of attention because it outperformed the previous state-of-the-art grid CRFs~\citep{shotton09textonboost} by a large margin on dense image labeling tasks, e.g., semantic image segmentation.
Naturally, there were many works that combined DenseCRF with CNN-based feature extractors to boost the performance even further.
In particular, \citet{zheng2015conditional,schwing15} used the joint training procedure, but \citet{chen2015deeplab,chen2018deeplab} used different variants of the stage procedure.
Because of complexity of tuning the parameters of DenseCRF, \citet{chen17deeplabv3} in their recent DeepLab-v3 system abandoned using DenseCRF at all.
\citet{vemulapalli2016gaussian,GAUSS,chandra2017dense} substituted the discrete DenseCRF model with the continuous Gaussian CRF, because it allowed relatively fast exact inference by solving a system of linear equations.
However, again due to the instabilities the joint training process (\citet{chandra2017dense} mentioned this explicitly), they used the stage training procedure.

Note that in many of the aforementioned works the authors used terminology ``joint end-to-end training'' to actually refer to the end-to-end fine-tuning of a system, some parts of which had been trained on the same data at an earlier stage.
For the purposes of our study, we refer to such procedures as stage training.
Also note that most of the aforementioned systems used some part of the networks pre-trained on different data, i.e., CNN feature extractors trained on ImageNet or word embeddings pre-trained on the Wikipedia corpus.
We chose not to consider such pre-training as a separate stage because from a practical point of view such pre-training is usually done separately from building a system for the task at hand, i.e., we just initialize the system from some pretrained model.

\textbf{Ill-conditioning.} Improper relative scaling, which we claim to be the main reason for the joint training to underperform, is closely related to ill-conditioning, which is a general problem in optimization and computational linear algebra.
Optimization literature offers a number of methods, mostly in the contexts of solving systems of linear and differential equations or convex optimization.
The most popular approach is to use preconditioning \citep{axelsson1985survey} and its 
special cases, e.g., variable rescaling and various regularization schemes.
These ideas have been successfully applied for training neural networks \citep{bottou2012stochastic} and laid the foundation for layer-wise normalization techniques, such as Batch-Normalization \citep{ioffe2015batch} and input normalization~\citep{krizhevsky2012imagenet}. However, in the context of structured prediction, the aforementioned techniques are less efficient since the potentials are objects of different types (according to our experiments layer-wise normalization cannot improve the relative scaling of potentials provided by different models).

\textbf{Signal aggregation.} The problem of aggregation signals of different kinds naturally arises in multimodal learning (see, e.g., \citep{baltruvsaitis2019multimodal}). For models of this type, the most common way of aggregation is a simple concatenation of signals (e.g., \citet{simonyan2014twoStream} constructed a state-of-the-art model for action recognition in video by concatenating the RGB and optical flow features computed by separate networks).
In such settings, it is natural to rescale different signals before aggregating them to reduce the discrepancy between their contributions.
However, in case of DSPMs, naive normalization of the potentials leads to the problems of joint training, which we are studying. 

To our knowledge, there is no general approach how to choose scaling of potentials in structured prediction models.
In this paper, we propose two approaches that can address the  issue and can be readily applied to most DSPMs.

\section{Models\label{sec:models}}

In this section, we briefly review the models that we use for our study together with the corresponding learning and prediction algorithms.

\subsection{Linear Chain Conditional Random Field} \label{lcrf}
The linear chain conditional random field (LCCRF) \citep{lafferty2001conditional} is one of the most standard models for sequence labeling tasks.
Given a pair of input and output variables $X,Y$ (the inputs and outputs consist of $L$~elements, i.e., $X=\{x_j\}_{j=1}^L$ $~{\text{and\ } Y = \{y_j\}_{j=1}^L}$, output variables $y_j$ take one of the $M$~values)
the LCCRF model is defined by its score function (a.k.a. negative energy function)
\begin{equation}
   F(Y\mid X, \theta) =  \sum\limits_{j=1}^L U_{\theta}(x_j, y_j) + \sum\limits_{j=1}^{L-1} W_{\theta}(y_j, y_{j+1}),
\end{equation}
where the unary potential $U_{\theta}(x_j, y_j)$ measures the confidence of assigning the label $y_j$ to the input $x_j$ and the pairwise potential $W_{\theta}(y_j, y_{j+1})$ scores the assignment of the pair of consecutive output variables $(y_{j}, y_{j+1})$. In this model, each output variable $y_j$ is directly influenced only by the corresponding input variable $x_j$ and the neighbouring output variables~$y_{j-1}$~and~$y_{j+1}$.

The LCCRF likelihood is defined by the normalized exponent of the score function
\begin{equation}\label{prob}
    P(Y\mid X,\theta) = \frac{1}{Z(\theta)}\exp\{F(Y\mid X, \theta)\},
\end{equation}
where $Z(\theta)$ denotes the partition function which ensures proper normalization of the probability function. Consequently, we can define the marginal distribution (which we call simply marginals) on the hidden variable $y_j$ as  the summation of $P(Y|X,\theta)$ over all possible states of all output variables except~$y_j$:
\begin{equation}
    p_j(y_j\mid X,\theta) = \sum\limits_{Y\setminus y_j}\frac{1}{Z(\theta)}\exp\{F(Y\mid X, \theta)\}.
\end{equation}
Given the parameters $\theta$ and input variables $X$ we can make prediction in several different ways, e.g., find the output variable assignment with the maximal score value, which coincides with the Maximum A Posteriori (MAP) estimate, or take the argmax of each marginal distribution.

To learn the parameters $\theta$ of LCCRF, we can use the maximum likelihood training, i.e., minimize the objective
\begin{equation}
    \mathcal{L}_{\text{MLE}}(X,Y \mid \theta) := -\log{P(Y\mid X,\theta)},
\end{equation}
which depends on the computation of the partition function~$Z(\theta)$.
An alternative choice for the training objective is the marginal likelihood~\cite{kakade2002} (equivalent to the standard cross-entropy loss defined on the marginals):
\begin{equation}\label{crosstrain}
    \mathcal{L}_{\text{CE}}(X,Y\mid \theta) := -\frac{1}{L} \sum\limits_{j=1}^L \log p_j(y_j \mid X,\theta).
\end{equation}
Instead of computing the marginals exactly, we can also approximate them, e.g., with the mean-field scheme, which in the case of LCCRF consists in the following updates:
\begin{align}
\log q_j(y_j) = \mathbb{E}_{q_{j+1}} W(y_j, y_{j+1}) &+ \mathbb{E}_{q_{j-1}} W(y_{j-1}, y_{j}) \nonumber\\
& + U(x_j,y_j) + C_j,
\end{align}
where $C_j$ is the normalization constant which can be computed in the linear in the number of labels~$M$ time.
For LCCRF, this approximation is not necessary since the exact computation of marginals is tractable, but for some more complicated models, such as DenseCRF \citep{krahenbuhl2011efficient}, the mean-field approximation is the only tractable scheme.

As an alternative to probabilistic training we can use the Structured Support Vector Machine, SSVM \citep{taskar2003max,tsochantaridis2005large}.
The SSVM objective seeks the parameters $\theta$ that push the score $F(Y\mid X,\theta)$ on the correct configuration $Y$ to be larger than the score $F(Y'\mid X,\theta)$ on other  labelings~$Y'$ with some margin denoted as $\Delta(Y,Y')$:
\begin{align}
    \mathcal{L}_{\text{SSVM}}(X,Y\mid\theta) := &\max_{Y'} \left\{F(Y'\mid X,\theta) + \Delta(Y,Y') \right\}\nonumber\\
    &- F(Y\mid X,\theta).
\end{align}
In this work, we consider the margin $\Delta(Y,Y')$ defined by the normalized Hamming distance between $Y$  and $Y'$
\begin{equation}
    \Delta(Y,Y') = \frac{1}{L} \sum\nolimits_{i=1}^L \mathbb{I}[y_i \neq y'_i],
\end{equation}
which is natural for sequence labeling tasks.

In the case of the SSVM and maximum likelihood training, we use the MAP inference during for prediction. In the case of the cross-entropy training, we take an argmax of each unary marginal.

In general, the computation of the partition function, marginals and  MAP estimate has complexity exponential in $L$ as it involves the summation or maximization over all possible configurations of $Y$.
However, the chain-like structure of the model allows to solve all these problems efficiently in linear in the number of entities $L$ time by dynamic programming algorithms also known as max-sum and sum-product message-passing \citep{Bishop:2006}.

\subsection{Gaussian Conditional Random Field}
In this section, we review a model for dense labeling of images, e.g., the image segmentation task, known as Gaussian Conditional Random Field, GCRF \citep{GAUSS,chandra2017dense}. 
The GCRF model is defined by the score function
\begin{equation}\label{GCRFenergy}
   F(s\mid X,\Theta) = -\frac{1}{2}s^T(W_{\theta}(X)+\lambda I)s + U_{\theta}(X)s, 
\end{equation}
where $W_{\theta}(X)$ is the symmetric positive semi-definite matrix of size $(LM)\times(LM)$ that corresponds to the pairwise potentials between pixel-label combinations and $U_{\theta}(X) \in \mathbb{R}^{LM}$ is the vector of unary terms.
Here $L$ stands for the number of pixels in the input image and $M$ for the number of possible pixel labels.
The vector $s$ consists of the stacked  scores for each instance and is continuous (unlike LCCRF):
\begin{equation}
    s = \left[s_{1,1},\dots,s_{1,L}|\dots|s_{M,1},\dots,s_{M,L}\right],
\end{equation}
where $s_{k,j}$ is the score for pixel $j$ to be labelled with label $k$.
The unary potentials $U_{\theta}(X)$ are of the structure similar to the score vector~$s$, i.e.,
\begin{equation}
    U_{\theta}(X) = \left[u_{\theta}^1(X)|\dots|u_{\theta}^M(X)\right],
\end{equation}
where the symbols $u_{\theta}^k(X)$ stand for the per-class unary potentials.

The $\lambda I$ term of equation~\eqref{GCRFenergy} with $\lambda > 0$ enforces the matrix of the quadratic term to be strictly positive-definite, which allows to compute the maximum point in a closed form
\begin{equation}
\label{eq:gcrf_solution}
    s^{*}(X,\theta) := (W_{\theta}(X)+\lambda I)^{-1}U_{\theta}(X).
\end{equation}
At the evaluation stage, the prediction is done by taking the instance-wise argmax of the score vector $s^{*}$
\begin{equation}
    \hat{Y} := [\argmax_k s^{*}_{k,1}, \dots, \argmax_k s^{*}_{k,L}].
\end{equation}
Hence, the inference task only involves solving a system of linear equations, which can be efficiently done for large matrices with the conjugate gradient iterative algorithm~\citep{shewchuk1994introduction}. 

Given the maximum point~\eqref{eq:gcrf_solution} we follow \citet{GAUSS} and define the marginal distribution of pixel~$j$ taking different labels by feeding the scores~$s^{*}_{:,j}(X,\theta)$ into the softmax function:
\begin{equation}\label{eq:softmaxmarginals}
p_j(y_j\mid X,\theta) = \text{softmax}_{y_j}\left(s^{*}_{:,j}(X,\theta)\right).
\end{equation}
Please note that the marginals~\eqref{eq:softmaxmarginals} are distributions on the discrete domain~$\{1,\dots,M\}$ and do not have any direct relationship to the actual marginal distributions of the Gaussian that corresponds to~\eqref{GCRFenergy}. 
Using~\eqref{eq:softmaxmarginals} we learn the parameters~$\theta$ by optimizing the cross-entropy objective~\eqref{crosstrain}.

To optimize over $W_{\theta}(X)$ and $U_{\theta}(X)$ terms parameterized by an arbitrary model, we need to obtain the gradients of the loss function $\mathcal{L}(X,Y\mid\theta)$ w.r.t.\ them. However, the direct propagation of gradients through the linear system solver is inefficient as it involves roll-out of the whole iterative procedure. Fortunately, in this case, the backward pass can be computed by solving another system of linear equations \citep{chandra2017dense}:
\begin{align} 
    \frac{\partial \mathcal{L}}{\partial U} = (W+\lambda I)^{-1}\frac{\partial  \mathcal{L}}{\partial s^{*}} \ , \ \ \ \ \ \ \ \ \frac{\partial \mathcal{L}}{\partial W} = - \frac{\partial \mathcal{L}}{\partial U} \otimes s^{*}, \label{eq:w}
\end{align}
where we omit the dependence of the potentials on $X$, $Y$ and $\theta$ for brevity and use $\otimes$ to denote the Kronecker product.
On the backward pass, we are given $\frac{\partial  \mathcal{L}}{\partial s^{*}}$, hence, in order to compute both $\frac{\partial \mathcal{L}}{\partial U}$ and $\frac{\partial \mathcal{L}}{\partial W}$ we reuse $s^{*}$ from the forward pass and solve the system of linear equations~\eqref{eq:w}.

We now define the potentials of the score~\eqref{GCRFenergy}.
The unary potentials are the outputs of a CNN applied to the input image in the fully-convolutional way.
We use the pairwise term~$W$ in the form of the class-agnostic Potts model \citep{chandra2017dense}, where the pairwise potential for the connected pixels to have different labels is computed as the dot-product of the pixel embeddings $\mathcal{A}_i$ (each pixel has the embedding  computed as the output of a CNN similar to the unaries).
In this setting, the matrix of pairwise potentials has the following form:
\begin{equation}
W = \begin{bmatrix}
    0 & \mathcal{A}\mathcal{A}^T\\
    \mathcal{A}\mathcal{A}^T & 0
\end{bmatrix}.
\end{equation}

\section{Scaling Methods\label{sec:scaling}}

\subsection{Online Scaling}\label{online}
We now introduce our first method to automatically choose a good relative scaling between the unary and pairwise potentials. Consider a score function $F(Y|X,\theta)$ as the function of the potentials $U_{\theta}(X)$ and $W_{\theta}(X)$, which are conditioned on the input and parameters:
\begin{equation}
    F(Y\mid X,\theta) = F(Y\mid U_{\theta}(X), W_{\theta}(X)).
\end{equation}
In addition, we assume that we are given a training objective, which also can be viewed as a function of $U_{\theta}(X)$ and $W_{\theta}(X)$ as it depends on the score function and ground-truth output variable $Y$. We denote  the training objective on a training pair $X,Y$ by~$\mathcal{L}(Y,U_{\theta}(X), W_{\theta}(X))$.

Now we define the scaling factor $\alpha$ that essentially represents the trade-off between the unary and pairwise components of the score function:
\begin{equation}
    \hat{F}_{\alpha}(Y\mid X,\theta) = F(Y\mid \alpha U_{\theta}(X), W_{\theta}(X)).
\end{equation}
Consequently, the training objective is modified as~$\mathcal{L}(Y,\alpha U_{\theta}(X), W_{\theta}(X))$. 
We denote the modified loss averaged over a subset $\mathcal{D}$ of objects by $\mathcal{L}_{\alpha}(\mathcal{D})$:
\begin{equation}
    \mathcal{L}_{\alpha}(\mathcal{D}) = \frac{1}{|\mathcal{D}|}\sum\limits_{(X,Y)\in\mathcal{D}} \mathcal{L}(Y,\alpha U_{\theta}(X), W_{\theta}(X))
\end{equation}
Now we introduce our online scaling algorithm. After each training epoch (i.e., a pass over the training set), we choose the scaling factor $\alpha$ via the grid search on the training set (subset for speed-up) $\mathcal{D}$ with respect to~$\mathcal{L}_{\alpha}(\mathcal{D})$. At the evaluation stage, we perform inference according to~$\hat{F}_{\alpha}(Y\mid X,\theta)$. Algorithm~\ref{onlinealgo} summarizes this approach.

\begin{algorithm}[htbp]
\begin{algorithmic}[1]
\STATE{\textbf{initialize} $\theta$}
\STATE{\text{epoch := 0}, $\alpha := 1$}
\FOR{\text{epoch} $<$ \text{n\_epoch}}
\FOR{$(X,Y)$ \text{in training set}}
\STATE{\text{compute} $(U_{\theta}(X), W_{\theta}(X))$}
\STATE{\text{compute} $\mathcal{L}(Y,\alpha U_{\theta}(X), W_{\theta}(X))$}
\STATE{\text{update} $\theta$ \text{via an SGD step}}
\ENDFOR
\STATE{\text{epoch += 1}}
\STATE{$\mathcal{D} :=$ \text{training set or its subset}}
\STATE{pick the best~$\alpha_{\text{best}}$ minimizing~$\mathcal{L}_{\alpha}(\mathcal{D})$ on an $\alpha$-grid} 
\STATE{$\alpha := \alpha_{\text{best}}$}
\ENDFOR
\end{algorithmic}
\caption{Online Scaling\label{onlinealgo}}
\end{algorithm}

\textbf{Remark.} When the unary potentials are defined as the output of a linear layer tweaking the constant $\alpha$ is equivalent to the special choice of initialization and learning rate for the parameters of the last linear layer when the stochastic optimization is done by the regular SGD (see Appendix~\ref{proof}). 

\subsection{Offline Scaling}

The online scaling technique has a computational bottleneck, which is the grid search w.r.t.\ the scaling factor $\alpha$ after each training epoch. Unfortunately, for many models described in section \ref{online} it is impossible to determine a constant scaling factor $\alpha$ that works well for the whole training procedure.
We will now parameterize the potentials to  approximately disentangle the potential individual norm and relative scaling between unary and pairwise terms.
We parameterize the potentials as follows:
\begin{align}
    &\hat{U}_{\theta}(X) := \alpha \frac{U_{\theta}(X)}{\|U_{\theta}(X)\|}, \ \ \ 
    \hat{W}_{\theta}(X) := \frac{W_{\theta}(X)}{\|W_{\theta}(X)\|},
\end{align}
for some fixed scaling factor $\alpha$, which can be tuned via validation.
Here, for a matrix $M \in \mathbb{R}^k\times \mathbb{R}^l$, we denote the averaged absolute value of its elements 
by $\|M\|$.

Consequently, the score function and the training objective are modified in the following way:
\begin{align*}
    &\hat{F}(Y\mid X,\theta) := F(Y\mid \hat{U}_{\theta}(X), \hat{W}_{\theta}(X))\\
    &\hat{\mathcal{L}}(Y,U_{\theta}(X), W_{\theta}(X)) := \mathcal{L}(Y,\hat{U}_{\theta}(X), \hat{W}_{\theta}(X))
\end{align*}
In addition to the direct modification of the score function, we can also enforce proper scaling implicitly via the regularization term added to the training objective:
\begin{align}
\mathcal{R}(U_{\theta}(X), W_{\theta}(X)) := \lambda \left(\frac{\|U_{\theta}(X)\|}{\|W_{\theta}(X)\|} - \alpha\right)^2
\end{align}
The hyperparameters $\lambda$ and $\alpha$ may be tuned on the validation set. The key difference of this approach from the previous one is that we do not modify the potentials directly.

\textbf{Remarks.} We found that the score function parameterization that rescales both the unary and pairwise potentials with a constant factor $\alpha$, i.e., changes the temperature, did not affect the model performance (see results in  Appendix~\ref{temperature}). In addition, the underperformance of the joint training is not related to the particular choice of the optimize. We report results for the Adam and SGD optimizers in Section~\ref{expsexps} and Appendix~\ref{sgd_study}, respectively.

\section{Experiments\label{sec:experiments}}
In this section, we evaluate the stage and joint training procedures together with the methods proposed in Section~\ref{sec:scaling} on the tree different tasks.
Section~\ref{sec:tasks} explains the tasks, models and evaluation methodology.
Section~\ref{expsexps} contains the experiment results and the discussion.

\subsection{Tasks}\label{sec:tasks}

\vspace{-0.1cm}\textbf{Optical Character Recognition.}\label{ocrdesc}
In the optical character recognition (OCR) task posed by \citet{taskar2003max},\footnote{\url{http://ai.stanford.edu/~btaskar/ocr/}}
we need to recognize an English word given images of its letters. All the first letters of the words in the dataset are cut off due to possible capitalization. 
The OCR task uses the averaged accuracy~$\frac{1}{N}\sum\nolimits_{i=1}^N\frac{1}{L_i}\sum\nolimits_{j=1}^{L_i} \mathbb{I}[\hat{y}_j^i = y_j^i]$ as the performance metric.
Here $\{\hat{y}^i_j\}_{j=1}^{L_i}$ is the prediction for word $i$ and $\{y^i_j\}_{j=1}^{L_i}$ is its ground-truth labeling.

The OCR dataset contains approximately 7000  words and is divided into 10 folds. We take one fold for validation purposes and use it only for selecting hyperparameters.
On the remaining 9 folds, we do the 9-fold cross-validation (train on 8 folds and test on 1, repeat 9 times).

For this task, we use the linear-chain CRF model. The unary potentials are obtained by applying the LeNet5-like \citep{lecun1998gradient} convolutional neural network to letter images (each image is zero padded to the size of $32 \times 32$). This unary CNN consists of the two convolutional blocks with 10 and 20 filters, respectively, and the kernel size of 5. Each convolution is followed by the ReLU activation \citep{dahl2013improving} and the $2\times 2$ max-pooling. 
The features obtained by the convolutional blocks are then passed through the two fully-connected layers with output sizes of 140 and 26, respectively, and the ReLU activation after the first fully-connected layer.
The pairwise potentials are parameterized with a real-valued matrix $W$ of size $26 \times 26$, where each matrix element represents the corresponding pairwise potential of the assignment of the neighboring variables~$(y_i, y_{i+1})$.

\textbf{CoNLL Chunking.}
Text chunking consists in dividing a text into syntactically correlated chunks of consecutive words. The CoNLL-2000 chunking dataset\footnote{\url{https://www.clips.uantwerpen.be/conll2000/chunking/}} \citep{chunk} contains 211727 tokens for training and 47377 tokens for the testing. Train and test data for this task are derived from the Wall Street Journal corpus (WSJ) and include words and part-of-speech (POS) tags as the input and chunking tags as the output (23 different chunking tags). For validation, we keep 20\%  of the training data.
The performance on this task is measured by the chunk $F_1$, which we compute with the official evaluation script.

We now review the system for which joint training works well. We closely follow the hierarchical recurrent network with CRF on top \citep{yang2016multi}, but substitute the BiGRU cells with BiLSTM. In more details, the unary predictor has the following hierarchical structure: each word is first encoded with a character-level BiLSTM and the obtained feature vector is concatenated with the SENNA embedding of the word \citep{collobert2011natural} and the POS tag embedding. After that, each word combined feature vector is passed through the word-level BiLSTM to obtain the unary potentials. And finally, the unary potentials are passed to the linear-chain CRF model with the pairwise potentials parameterized by a matrix of size $23 \times 23$. The character-level LSTM has 2 layers with the state of size 50, the word-level LSTM has 2 layers with the state of size 300. The size of the letter embedding (encodes each letter for the character-level LSTM) equals 10, likewise the POS-tag embedding. Each SENNA embedding has the size of 50.
We add a learnable embedding which corresponds to the words that did not appear in SENNA embeddings.

The table below reports results of the stage and joint training procedures (three standard training objectives).
This table shows that joint training works well as is.

\setlength\extrarowheight{1pt}
\begin{center}
\vspace{-0.1cm}
\begin{tabular}{c|c|c}
\textbf{OBJECTIVE}& \textbf{STAGE}& \textbf{JOINT} \\
\hline
Cross-entropy &   94.7 & 94.9 \\
Structured SVM & 94.6 & 94.8 \\
Log-likelihood & 94.7  & 94.8 \\
\end{tabular}
\end{center}

\vspace{-0.1cm}We believe that the word-level BiLSTM cells in the model normalize the potentials properly so the scaling problem is not present.
If we substitute the word-level BiLSTM with the two linear layers (the ReLU activation in-between) the scaling breaks and the joint training starts to underperform.
We use this model for experiments in Section \ref{expsexps}.

{\renewcommand{\arraystretch}{1.2} 
\begin{table*}[htbp]
\definecolor{Gray}{gray}{0.9}
\begin{center}
\resizebox{\textwidth}{!}{
\begin{tabular}{@{}c|c|c|c|c|c|c@{}}
\multirow{2}{*}{\textbf{TASK}} & \multirow{2}{*}{\textbf{OBJECTIVE}}& \multirow{2}{*}{\textbf{STAGE}} & \multirow{2}{*}{\textbf{JOINT}} & \multirow{2}{*}{\textbf{ONLINE}} & \multirow{2}{*}{\textbf{OFFLINE}} & \multirow{2}{*}{\shortstack{\textbf{OFFLINE}\\\textbf{(Reg)}}} \\[-1mm]
& & & & & &   \\
\hline
\hline
\multirow{4}{*}{\textbf{OCR}} & Cross-entropy  &  97.2 $\pm$ 0.1 , 41& \cellcolor[HTML]{FFF3F5}96.5 $\pm$ 0.2 , 23 & 97.3 $\pm$ 0.1 , 27 & \cellcolor[HTML]{F0F8FF}97.2 $\pm$ 0.1 , 23 &\cellcolor[HTML]{F0F8FF} 97.1 $\pm$ 0.2 , 23 \\
& Structured SVM  & 97.0 $\pm$ 0.3 , 43  & \cellcolor[HTML]{FFF3F5}96.4 $\pm$ 0.4 , 23 & 97.0 $\pm$ 0.2  , 27&\cellcolor[HTML]{F0F8FF} 97.0 $\pm$ 0.3 , 23 &\cellcolor[HTML]{F0F8FF} 96.9 $\pm$ 0.4 , 23  \\
& Log-likelihood  & 97.2 $\pm$ 0.1 , 42 & \cellcolor[HTML]{FFF3F5}96.5 $\pm$ 0.3 , 23 & 97.2 $\pm$ 0.1 , 27 &\cellcolor[HTML]{F0F8FF} 97.1 $\pm$ 0.1 , 23 &\cellcolor[HTML]{F0F8FF} 97.1 $\pm$ 0.2 , 23 \\
& Cross-entropy (MF)  & 97.1 $\pm$ 0.1 , 50 & \cellcolor[HTML]{FFF3F5}96.6 $\pm$ 0.2 , 26 & 97.1 $\pm$ 0.1 , 30 &\cellcolor[HTML]{F0F8FF} 97.1 $\pm$ 0.1 , 26 &\cellcolor[HTML]{F0F8FF} 97.0 $\pm$ 0.1 , 27 \\
\hline
\multirow{4}{*}{\textbf{Chunking}} & Cross-entropy  & 89.5 $\pm$ 0.3 , 94 & \cellcolor[HTML]{FFF3F5}87.9 $\pm$ 0.4 , 51 & 89.6 $\pm$ 0.3 , 55 &\cellcolor[HTML]{F0F8FF} 89.5 $\pm$ 0.3 , 53 &\cellcolor[HTML]{F0F8FF} 89.5  $\pm$ 0.3 , 53\\
& Structured SVM  & 89.2 $\pm$ 0.5 , 97 & \cellcolor[HTML]{FFF3F5}87.6 $\pm$ 0.6 , 54 &  89.3 $\pm$  0.4 , 57 &\cellcolor[HTML]{F0F8FF}  89.2 $\pm$ 0.5 , 56 &\cellcolor[HTML]{F0F8FF} 89.1 $ \pm$ 0.6 , 56 \\
& Log-likelihood  & 89.4 $\pm$ 0.3 , 95 & \cellcolor[HTML]{FFF3F5}87.9 $\pm$ 0.4 , 53 & 89.5 $\pm$ 0.3 , 56  &\cellcolor[HTML]{F0F8FF} 89.5 $\pm$ 0.3 , 54 &\cellcolor[HTML]{F0F8FF} 89.4 $\pm$ 0.4 , 54\\
& Cross-entropy (MF)  & 89.3 $\pm$ 0.3 , 104  & \cellcolor[HTML]{FFF3F5}87.7 $\pm$ 0.3 , 61 & 89.3 $\pm$ 0.2 , 66 &\cellcolor[HTML]{F0F8FF} 89.3 $\pm$ 0.3 , 64 &\cellcolor[HTML]{F0F8FF} 89.2 $\pm$ 0.3 , 64 \\
\hline
\textbf{Bin. segm.} & Cross-entropy & 86.5 $\pm$ 0.2 , 80  & \cellcolor[HTML]{FFF3F5}85.6 $\pm$ 0.3 , 44 & 86.6 $\pm$ 0.2 , 45  &\cellcolor[HTML]{F0F8FF} 86.6 $\pm$ 0.2 , 44 &\cellcolor[HTML]{F0F8FF} 86.5 $\pm$ 0.3 , 44\\[-1mm]
\end{tabular}
}
\end{center}
\caption{Results of the main experiment.
We report the performance (metrics are task specific, see Section~\ref{sec:tasks}) and the mean training time (in minutes) of the stage and joint training procedures together with the three proposed ways of fixing the joint training.
On the three tasks (OCR, chunking and binary segmentation), we studied the systems trained with the suitable objectives.
For all the performance metrics, we report the mean and standard deviation w.r.t.\ 8 random seeds.
\label{summary}
}
\vspace{-3mm}
\end{table*}
}

\textbf{Binary Segmentation.}
We consider the Weizmann Horses dataset \citep{borenstein2002class} for the binary segmentation task. The task consists in classifying image pixels into foreground or background. The dataset consists of 328 side-view color images of horses that were manually segmented. We keep 50 samples from the training set for validation. We randomly divide the remaining dataset into 5 almost equal folds and do 5-fold cross validation.
For the performance measure, this task uses the intersection over union scores (Jaccard index) between pixel labelings.

For the binary segmentation task, we use the GCRF model. The unary network is a UNet-like \citep{ronneberger2015u} fully-convolutional neural network which is a typical encoder-decoder model. The encoder consists of convolutional blocks, which are basically a stack of convolutions, batch normalization and ReLU activations. After each convolutional block, we downsample with max-pooling. The decoder consist of similar convolutional blocks and upsampling is done by the bilinear interpolation. The shortcut connections between the convolutional blocks of the encoder and decoder of the identical resolution allow to use the features of different resolutions for more robust segmentation. The architecture of pairwise network for pixel embeddings is similar to the unary network, but without the regression layer at the end.

\subsection{Results and Discussion}\label{expsexps}
Table~\ref{summary} reports the results of different training schemes (stage and regular joint training with~$\alpha=1$; the three methods to improve the joint training described in Section~\ref{sec:scaling}) on the three tasks described in Section~\ref{sec:tasks}.
On the OCR and chunking tasks, we also compare the four training objectives: the cross-entropy on the exact marginals, structured SVM, log-likelihood and the cross-entropy on the marginals estimated with the mean-field (MF) approximation.
For binary segmentation, we use the only objective available, which is the cross-entropy on the marginals estimated by the softmax of~\eqref{eq:softmaxmarginals}.
We always report the mean and standard deviation computed over 8 random seeds and the training time.
The details on the hyperparameter choice and stopping criterion (consequently, the time measurements) are provided in Appendix~\ref{experiment_details}.
As another baseline, the table below shows the performance of the unary predictors only, which is significantly worse  than the performance of the full models in all tasks.
\vspace{-0.5em}
\begin{center}
\begin{tabular}{@{}c|c|c|c@{}}
& \textbf{OCR} & \textbf{CHUNKING} & \textbf{BIN. SEGM.} \\
\hline
\textbf{Unary} & 91.8 $\pm$ 0.2  & 86.2 $\pm$ 0.4  & 84.1 $\pm$ 0.2\\
\end{tabular}
\end{center}

\textbf{Discussion.} Table~\ref{summary} shows that the proposed scaling techniques perform up to noise identically to the stage training procedure (but are two times faster, and have less hyperparameters to tune, see Appendix~\ref{experiment_details} for the details) and are always superior to the joint training.
The online scaling scheme~\ref{online} has the most stable training at the cost of additional grid search steps, which increases the training time.
The offline scaling approaches allow to reduce computational complexity at the cost of having more hyperparameters.
The regularization approach has slightly higher variance and has one extra hyperparameter, which increases the tuning time.

To strengthen our conclusions, we run additional experiments to investigate possible confounding factors of the effects reported in Table~\ref{summary}.
All the models of Table~\ref{summary} were trained with the Adam optimizer~\cite{Adam}, but we have tried alternatives and observed similar story, except that all the methods performed worse. We report the full results for SGD with momentum in Appendix~\ref{sgd_study}.
As another possible confounding factor we investigate the joint scaling of the potentials (a.k.a. the temperature), which implies multiplying both unary and pairwise potentials on the same constant~$\alpha$.
Our experiments with temperature scaling showed that it did not solve underperformance of the joint training at all (see Appendix~\ref{temperature} for the results).
We would also like to explicitly mention that the problems of the joint training are not related to over-fitting, which is confirmed by the fact that the same discrepancy between joint and stage procedures appears in the training error as well as in the test error, which we report in all our experiments.

\section{Conclusion\label{sec:conclusion}}
In this paper, we study the situations when the joint training of the deep structured-prediction energy-based models unexpectedly performs worse than the inconvenient multistage training approach.
We conjecture that the source of the problem lies in the improper relative scaling of the summands of the energy (that by construction are of different nature) and propose several ways (with different engineering properties) to fix the problem.
We apply the proposed techniques to the two well-established models (linear-chain CRF and Gaussian CRF) on three different tasks and demonstrate that they indeed improve performance of the joint training to the level of the stage training procedure.

\section*{Acknowledgements}
This work was partly supported by Samsung Research, Samsung Electronics. 

{
\bibliography{refs}
}

\newpage
\appendix

\twocolumn[
\icmltitle{Supplementary Material (Appendix) \\[2mm]
Scaling Matters in Deep Structured-Prediction Models}
]

\section{Additional Experiments}

\subsection{Main Experiment with Another Optimizer}
\label{sgd_study}

To show whether the poor performance of the basic joint training procedure and applicability of our approaches are not optimizer dependent, we repeat the main experiments of Table~\ref{summary} with a different  optimization method~-- SGD with momentum.
We report the results in Table~\ref{summary_SGD}.
As it can be seen the overall behavior stays the same, but the performance of all methods with the SGD optimizer is slightly worse than with Adam~(Table~\ref{summary}). This observation suggests that the reason of joint procedure to underperform is not related to a particular chose of optimization algorithm (we have also tried RMSprop, but did not run a larger scale experiments with it). As the performance of  SGD  is  worse compared to Adam we have chosen Adam for the main experiment reported in Section~\ref{expsexps}.

\subsection{Main Experiment, but with Temperature Tuned}
\label{temperature}

To show that the issue of joint training is not related to the joint scaling of the potentials, we consider a model with the extra parameter~$\alpha$ in front of both unary and pairwise potentials:
\begin{align*}
    &\hat{U}_{\theta}(X) :=   \alpha  U_{\theta}(X), \ \ \ 
    \hat{W}_{\theta}(X) :=   \alpha  W_{\theta}(X).
\end{align*}
The scaling factor $\alpha$ is then chosen on the validation set. In this setting, we use the Adam optimizer. Table~\ref{summary_temp} shows that the joint scaling of the potentials does not significantly affect the resulting performance of the joint training procedure. Another scheme with  the ``hard'' global scaling parameterization
\begin{align*}
    &\hat{U}_{\theta}(X) :=   \alpha  \frac{U_{\theta}(X)}{\|U_{\theta}(X)\|}, \ \ \ 
    \hat{W}_{\theta}(X) :=   \alpha  \frac{W_{\theta}(X)}{\|W_{\theta}(X)\|}
\end{align*}
performs up to noise identically to the one reported in Table~\ref{summary_temp}.
These results allow us to conclude that the choice of the joint scaling (the temperature)  is not the main reason for the joint training to underperform.

\section{Hyperparameter Selection}
\label{experiment_details}

In this section, we discuss the hyperparameters of all used schemes. The most important hyperparameters for all procedures appeared to be  the learning rate and the number of training iterations. Consequently, the number of hyperparameters for stage procedure is 3 times larger than for the joint scheme (because we need to select the learning rate and the number of training iterations for each stage separately). For the proposed scaling methods, we also have additional parameters to tune, which we describe below. 

\textbf{Learning Rate and Stopping Criterion.} We now discuss the methodology of choosing the learning rate and the number of optimizer iterations in the cases of stage and joint procedures. For joint procedures, the scheme is straight-forward and does not differ from the regular deep learning pipeline. We choose the learning rate to ensure faster initial convergence (but without optimization diverging) and additionally reduce it by a factor of 10 after a number of epochs (3 for OCR, 1 for chunking, 5 for binary segmentation) with the dataset specific metric on the validation set changing less than $10^{-3}$ in the absolute value (we use  the \texttt{ReduceLROnPlateau} function implemented in PyTorch). The number of optimizer iterations is chosen in a similar way, i.e., if within enough epochs (7 for OCR, 3 for chunking, 13 for binary segmentation) the changes in the dataset specific metric on the validation set are smaller than   $10^{-3}$. In the case of stage training, the same procedure is repeated for each stage (three times) and the overall model performance is more sensitive to the selection of both the learning rate and number of iterations. When using these schemes, the joint training converged significantly faster in comparison to the stage scheme.

We would also like to mention that for the stage training it is sometimes beneficial not to train unary network to convergence, which makes choosing the stopping criterion for that stage much harder. The joint procedure does not have this issue. 

\textbf{Additional hyperparameters.} For the online scaling method, we need to set the grid to search for $\alpha$, the size of the subset to choose~$\alpha$ and the initial scale.
We observed that the uniform in the log scale grid defined as $\{2^{t} \mid t = -8, \dots, 8 \}$ worked well for all tasks we considered and there was no need to tune it.
For the subset size, it was sufficient to always use around a third of the training data (which was 2000, 3000 and 100 examples for OCR, chunking and binary segmentation, respectively). The initialization of~$\alpha$ also appeared not to be a sensitive parameter and it was sufficient to simply start with $\alpha=1$. 

For the offline methods, we had to set scaling~$\alpha$ and regularization weight~$\lambda$.
We had to select these hyperparameters by looking at the trained model results on the validation set. In this case, the complexity of tuning these parameters is the same as, e.g., choosing the weight decay parameter.

\section{Rescaling Equivalence}
\label{proof}
When the unary potentials are defined as the output of a linear layer $U_{\theta}(X) = V\phi_{\theta}(X)$ (the bias term is omitted for brevity) tweaking the constant $\alpha$ is equivalent to the special initialization and learning rate for the parameters $V$ when the stochastic optimization is done by the regular SGD. 
For brevity, we shortcut the loss function to $\mathcal{L}(U_{\theta}(X))$ omitting the dependence on $Y$ and the pairwise potentials. We now consider the two training settings: with scaling $\alpha$ and with the $\alpha$ times larger initialization for the parameter $V$: $
    V^{(0)}_{\text{i}} = \alpha V^{(0)}_{\text{s}},
$
where $V_{\text{i}}$ and $V_{\text{s}}$ correspond to the initialization and scaling settings, respectively, and the upper index defines the SGD step index.

Consider the SGD updates with the learning rate $\eta$ for the parameter $V$ and the gradients of the loss function $\mathcal{L}(U_{\theta}(X))$ w.r.t. $\phi_{\theta}(X)$. In the first setting, we have:
\begin{align*}
    &V^{(1)}_{{\text{s}}} = V^{(0)}_{\text{s}} - \eta \nabla \mathcal{L}\left(\alpha V^{(0)}_{\text{s}}\phi_{\theta}(X)\right)\alpha\phi^T_{\theta}(X),\\
    &\frac{\partial \mathcal{L}}{\partial\phi_{\theta}}(X) = \left(\alpha V^{(0)}_{\text{s}}\right)^{T} \nabla \mathcal{L}\left(\alpha V^{(0)}_{\text{s}}\phi_{\theta}(X)\right)
\end{align*}
In the second setting with the $\eta^{*} := \alpha^2\eta$, we have:
\begin{align*}
    &V^{(1)}_{\text{i}} = V^{(0)}_{\text{i}} - \eta^{*} \nabla \mathcal{L}\left( V^{(0)}_{\text{i}}\phi_{\theta}(X)\right)\phi^T_{\theta}(X),\\
    &\frac{\partial \mathcal{L}}{\partial\phi_{\theta}}(X) = \left(V^{(0)}_{\text{i}}\right)^{T} \nabla \mathcal{L}\left(V^{(0)}_{\text{i}}\phi_{\theta}(X)\right),
\end{align*}
which gives
$
V^{(1)}_{\text{i}} = \alpha V^{(1)}_{\text{s}}
$ 
and equal partial derivatives of the loss function w.r.t. $\phi_{\theta}$. By induction, it can be easily seen that after $n$ iterations of SGD we have
 $
V^{(n)}_{\text{i}} = \alpha V^{(n)}_{\text{s}}
$ 
and the partial derivatives of the loss function w.r.t. $\phi_{\theta}$ are also equal. Hence, the two training runs are equivalent.

{\renewcommand{\arraystretch}{1.2} 
\begin{table*}[htbp]
\begin{center}
\begin{tabular}{@{}c|c|c|c|c|c|c@{}}
\multirow{2}{*}{\textbf{TASK}} & \multirow{2}{*}{\textbf{OBJECTIVE}}& \multirow{2}{*}{\textbf{STAGE}} & \multirow{2}{*}{\textbf{JOINT}} & \multirow{2}{*}{\textbf{ONLINE}} & \multirow{2}{*}{\textbf{OFFLINE}} & \multirow{2}{*}{\shortstack{\textbf{OFFLINE}\\\textbf{(Reg)}}} \\[-1mm]
& & & & & &   \\
\hline
\hline
\multirow{4}{*}{\textbf{OCR}} & Cross-entropy  &  97.1 $\pm$ 0.2& 96.4 $\pm$ 0.3 & 97.2 $\pm$ 0.1 & 97.2 $\pm$ 0.2 & 97.1 $\pm$ 0.2 \\
& Structured SVM  & 96.9 $\pm$ 0.3  & 96.2 $\pm$ 0.5 & 97.0 $\pm$ 0.3 & 96.9 $\pm$ 0.3 & 96.9 $\pm$ 0.4  \\
& Log-likelihood  & 97.1 $\pm$ 0.2 & 96.3 $\pm$ 0.3 & 97.1 $\pm$ 0.2 & 97.1 $\pm$ 0.2 & 97.0 $\pm$ 0.3  \\
& Cross-entropy (MF)  & 97.0 $\pm$ 0.1 & 96.5 $\pm$ 0.2  & 97.1 $\pm$ 0.1 & 97.1 $\pm$ 0.1 & 97.0 $\pm$ 0.2 \\
\hline
\multirow{4}{*}{\textbf{Chunking}} & Cross-entropy  & 89.5 $\pm$ 0.3 & 87.9 $\pm$ 0.4 & 89.5 $\pm$ 0.3 & 89.5 $\pm$ 0.3 & 89.4  $\pm$ 0.4\\
& Structured SVM  & 89.2 $\pm$ 0.5 & 87.5 $\pm$ 0.6 &  89.2 $\pm$  0.5  &  89.1 $\pm$ 0.6 & 89.1 $ \pm$ 0.6 \\
& Log-likelihood  & 89.4 $\pm$ 0.4 & 87.8 $\pm$ 0.5 & 89.5 $\pm$ 0.3  & 89.4 $\pm$ 0.3 & 89.3 $\pm$ 0.4\\
& Cross-entropy (MF)  & 89.2 $\pm$ 0.3  & 87.6 $\pm$ 0.3 & 89.3 $\pm$ 0.3 & 89.2 $\pm$ 0.3 & 89.1 $\pm$ 0.3 \\
\hline
\textbf{Bin. segm.} & Cross-entropy & 86.5 $\pm$ 0.2  & 85.5 $\pm$ 0.4  & 86.5 $\pm$ 0.2  & 86.5 $\pm$ 0.2 & 86.4 $\pm$ 0.3\\[-1mm]
\end{tabular}
\end{center}
\caption{Results of the main experiment repeated with the SGD optimizer.
\label{summary_SGD}
}
\vspace{-4mm}
\end{table*}
}
{\renewcommand{\arraystretch}{1.2} 
\begin{table*}[htbp]

\begin{center}
\begin{tabular}{@{}c|c|c|c|c@{}}
 {\textbf{TASK}} & {\textbf{OBJECTIVE}}&{\textbf{STAGE}} & {\textbf{JOINT}} & {\textbf{\textbf{TEMP}}} \\
\hline
\hline
\multirow{4}{*}{\textbf{OCR}} & Cross-entropy  &  97.2 $\pm$ 0.1& 96.5 $\pm$ 0.2 & 96.5 $\pm$ 0.2  \\
 & Structured SVM & 97.0 $\pm$ 0.3  & 96.4 $\pm$ 0.4 & 96.5 $\pm$ 0.3  \\
 & Log-likelihood  &97.2 $\pm$ 0.1 & 96.5 $\pm$ 0.3 & 96.5 $\pm$ 0.2  \\
 & Cross-entropy (MF) &97.1 $\pm$ 0.1 & 96.6 $\pm$ 0.2  & 96.7 $\pm$ 0.1 \\
\hline
 \multirow{4}{*}{\textbf{Chunking}} & Cross-entropy  &89.5 $\pm$ 0.3 & 87.9 $\pm$ 0.4 & 87.9 $\pm$ 0.3\\
 & Structured SVM  &89.2 $\pm$ 0.5 & 87.6 $\pm$ 0.6 &  87.7 $\pm$  0.5 \\
 & Log-likelihood  &89.4 $\pm$ 0.3 & 87.9 $\pm$ 0.4 & 87.9 $\pm$ 0.3\\
 & Cross-entropy (MF)  &89.3 $\pm$ 0.3  & 87.7 $\pm$ 0.3 & 87.8 $\pm$ 0.3\\
\hline
 \textbf{Bin. segm.} & Cross-entropy & 86.5 $\pm$ 0.2  & 85.6 $\pm$ 0.3  & 85.7 $\pm$ 0.3\\[-1mm]
\end{tabular}
\end{center}
\caption{Results of the main experiment, but with the temperature tuned. TEMP denotes the scheme with the global parameter~$\alpha$ described in Section~\ref{temperature}.
\label{summary_temp}
}
\vspace{-4mm}
\end{table*}
}

\end{document}